\documentclass[10pt,letterpaper,oneside,twocolumn]{IEEEtran}

\usepackage{graphicx}
\usepackage{amsfonts,amsmath,amssymb}
\usepackage{epsfig}
\usepackage{url}
\usepackage{cite}
\usepackage{multirow}
\usepackage{mathptmx} 
\usepackage{times}

\IEEEoverridecommandlockouts                              
\title{\LARGE \bf
Automated Epilepsy Diagnosis Using Interictal Scalp EEG}

\renewcommand{\baselinestretch}{1}

\author{Forrest Sheng Bao, Jue-Ming Gao, Jing Hu, Donald Y.-C. Lie, Yuanlin Zhang and K. J. Oommen
\thanks{This research is supported by the Fall 2007 Research Enrichment Fund Grant of Texas Tech University. Human subject data used in this research has been approved and are already exempt by Protection of Human Subjects Committee IRB  committee of Texas Tech University under ``501209 Deagnosis, Monitoring, Seizures Prediction and Intervention for Epilepsy Patients Using an Intelligent Scalp-EEG Signal Analysis System.''}
\thanks{F. S. Bao, D. Y.-C. Lie and Y. Zhang are with Dept. of Computer Science and Dept. of Electrical \& Computer Engineering, Texas Tech University, Lubbock 79409 Texas, USA
        {\tt\small forrest.bao@gmail.com}}%
\thanks{J. Hu and J.-M. Gao are with Dept. of Neurosurgery, Jiangsu Provincial Hospital of Chinese Medicine, Nanjing, Jiangsu China}%
\thanks{K. J. Oommen is with Dept. of Neurology, Texas Tech University Health Sciences Center, Lubbock 79409 Texas, USA}
}

\begin{document}

\maketitle
\thispagestyle{empty}
\pagestyle{empty}

\begin{abstract}
Approximately over 50 million people worldwide suffer from epilepsy. Traditional diagnosis of epilepsy relies on tedious visual screening by highly trained clinicians from lengthy EEG recording that contains the presence of seizure (ictal) activities. Nowadays, there are many automatic systems that can recognize seizure-related EEG signals to help the diagnosis. However, it is very costly and inconvenient to obtain long-term EEG data with seizure activities, especially in areas short of medical resources. We demonstrate in this paper that we can use the interictal scalp EEG data, which is much easier to collect than the ictal data, to automatically diagnose whether a person is epileptic. In our automated EEG recognition system, we extract three classes of features from the EEG data and build Probabilistic Neural Networks (PNNs) fed with these features. We optimize the feature extraction parameters and combine these PNNs through a voting mechanism. As a result, our system achieves an impressive 94.07\% accuracy, which is very close to reported human recognition accuracy by experienced medical professionals.
\end{abstract}

\section{Introduction}

\IEEEPARstart{E}{pilepsy} is the second most common neurological disorder, affecting 1\% of world population~\cite{BME_Prediction}. Eighty-five percent of patients with epilepsy live in the developing countries~\cite{WHO_Atlas}. Electroencephalogram (EEG) is routinely used clinically to diagnose epilepsy~\cite{purpose_of_EEG}. Long-term video-EEG monitoring can provide 90\% positive diagnostic information~\cite{90Positive} and it has become the golden standard in epilepsy diagnosis. For the purpose of this research, we define the term ``the diagnosis of epilepsy'' as the determination of whether a person is epileptic or non-epileptic~\cite{textbook}.

Traditional diagnostic methods rely on experts to visually inspect lengthy EEG recordings, which is time consuming and problematic due to the lack of clear differences in EEG activity between epileptic and non-epileptic seizures~\cite{Bigan_1998}, particularly in seizures of frontal origin. Many automated seizure recognition techniques, therefore, have emerged~\cite{One-Class_Novelty_Detection_for_Seizure_Analysis_from_Intracranial_EEG,1084730,243992,1028922,A_neural-network-based_detection_of_epilepsy,lvq,4167902,  Geva_1998, Bigan_1998, Alkan2005167, gorgan99,gotman1999}. 
The approach of using automatic seizure recognition/detection algorithms would still require the recording of clinical seizures. Therefore, very long continuous EEG recording, preferably with synchronized video for several days or weeks, are needed to capture the seizures. The long-term EEG recording can greatly disturb patients' daily lives. Another clinical concern is that very unfortunately, 50-75\% of epilepsy patients in the world reside in areas which lack the medical resources and trained clinicians, that are needed to make such a process feasible~\cite{WHO_Atlas}. Consequently, an automated EEG epilepsy diagnostic system would be very valuable if it does not require data containing seizure activities (i.e., ictal) to arrive at the diagnosis. However, to the authors' best knowledge, we are not aware of any report on automated epilepsy diagnostic system using only interictal \textit{scalp EEG} data.

Previous research has also attempted at creating automated epilepsy diagnostic systems using interictal EEG data~\cite{ICTAI2008, 4167902}. However, in those trials, only \textit{intracranial} EEG data from patients are used, and the EEG artifacts have been carefully removed manually. It is very expensive to obtain intracranial EEG recordings that are relatively artifact free for every epilepsy patient, which is especially impractical in poor and rural areas. Therefore, we have built an automated epilepsy diagnostic system with very good accuracy that can work with scalp EEG data that contain noise and artifacts.

Artificial Neural Network (ANN) has been used for seizure-related EEG recognition~\cite{4167902,lvq,A_neural-network-based_detection_of_epilepsy,1028922,Geva_1998, Alkan2005167}.We use in this work one kind of ANN as the classifier, namely the Probabilistic Neural Network (PNN)
, for its high speed, high accuracy and real-time property in updating network structure~\cite{PNN}.
It is very difficult to directly use raw EEG data as the input of an ANN~\cite{NTUpaper}. Therefore, the key is to parameterize the EEG data into features prior to the input into the ANN. We use features that are used in previous studies on seizure-related EEG, namely, the power spectral feature, fractal dimensions and Hjorth parameters. A simple classifier voting scheme~\cite{Pattern_Classification} and parameter optimization are used to improve the accuracy. 

\setlength{\unitlength}{0.4cm}
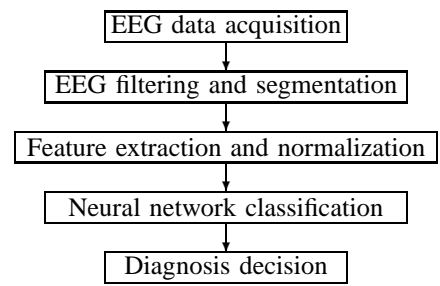
\begin{figure}[!hbt]
\begin{center}
 \begin{picture}(6,8.5)
\put(-1,0){\framebox(8,1){Diagnosis decision}}
\put(3,2){\vector(0,-1){1}}
\put(-3,2){\framebox(12,1){Neural network classification}}
\put(3,4){\vector(0,-1){1}}
\put(-4,4){\framebox(14,1){Feature extraction and normalization}}
\put(3,6){\vector(0,-1){1}}
\put(-3,6){\framebox(12,1){EEG filtering and segmentation}}
\put(3,8){\vector(0,-1){1}}
\put(-1,8){\framebox(8,1){EEG data acquisition}}
\end{picture}
\caption[fig]{Flow diagram of our EEG classification
scheme\label{diagram}}
\end{center}
\end{figure}

The final accuracy of our system on distinguishing interictal scalp EEG of epileptic patients vs. the scalp EEG of healthy people is 94.07\%, which is very close to currently reported human diagnosis accuracy~\cite{EEG_accuracy}.  

\section{Data Acquisition}
\label{data}

We compose a data set based on 22-channel routine scalp EEG recordings from Dept. of Neurology, Jiangsu Provincial Hospital of Chinese Medicine, China. 
The data is from 6 normal people and 6 epileptic patients (in interictal period only). It is recorded at 200 Hz sampling rate using the standard international 10-20 system with referential montage.  Whereas other research~\cite{4167902} EEG recordings are cut into segments of 4096 (i.e., $2^{12}$), our complete data set has $22,353$ segments per channel, and $491,766$ segments in total.


\section{Feature Extraction}
\label{feature_extraction}
Three classes of features are extracted to characterize EEG signal: Power Spectral Features, describing its energy distribution in the frequency domain, Fractal Dimensions outlining its fractal property, and Hjorth Parameters, modeling its chaotic behavior. 

\subsection{Power Spectral Features}

\begin{figure}[!hbt]
\begin{center}
\includegraphics[scale=0.26]{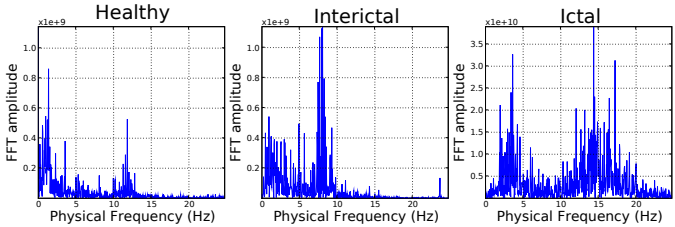}
\caption[fig]{Typical FFT results of 3 EEG segments (Raw data in $\mu$V)}
\label{FFT}
\end{center}
\end{figure}

As one can see from Fig. \ref{FFT}, power spectrum is a good way to distinguish different kinds of EEG signals.

To a time series $x_{1}, x_{2}, \cdots, x_{N}$, its Fast Fourier Transform (FFT) $X_{1}, X_{2}, \cdots, X_{N}$ is estimated as 
$$
X_{k} = \sum_{1}^{N}x_{n}W_{N}^{kn},\ \ k = 1,2,\cdots,N
$$
, where $W_{N}^{kn}= e^{\frac{-j2\pi kn}{N}}$ and $N$ is the series length.

Based on the FFT result, Power Spectral Intensity (PSI) of each $f_\text{step}$Hz bin in a given band $f_\text{low}$-$f_\text{up}$Hz is evaluated as 
\begin{equation}PSI_{k}=\sum_{i=\lfloor N\frac{f_\text{min}}{f_\text{s}}\rfloor}^{\lfloor N\frac{f_\text{max}}{f_\text{s}}\rfloor}X_{i}, \ \ \  k=1,2, \cdots, K
\label{PSI}
\end{equation}
, where $f_\text{min}=2k$, $f_\text{max}=2k+2$, $K=(f_\text{up}-f_\text{low})/{f_\text{step}}$, $f_\text{s}$ is the sampling rate and $N$ is the series length. $f_\text{min}$ and $f_\text{max}$ are the lower and upper boundaries of each bin, respectively.

We use  Relative Intensity Ratio (RIR) as the Power Spectral Features. It is defined as 
$$RIR_{j}=\frac{PSI_{j}}{\sum_{k=1}^{K} PSI_{k}}, \ \ \ j=1,2,\cdots, {(f_\text{up}-f_\text{low})}/{f_\text{step}}.$$

\subsection{Petrosian Fractal Dimension (PFD)}
PFD is defined as:
$$
\text{PFD} = \frac{\log_{10}{N}}{\log_{10}{N} + \log_{10}(\frac{N}{n+0.4N_{\delta}})}
$$
, where $N$ is the series length and $N_{\delta}$ is the number of sign changes in the signal derivative~\cite{Petrosian}.

\subsection{Higuchi Fractal Dimension (HFD)} 
Higuchi's algorithm~\cite{Higuchi} constructs $k$ new series from the original series $ x_{1}, x_{2}, \cdots, x_{N}$ by 
\begin{equation}
x_{m},x_{m+k},x_{m+2k},\cdots, x_{m+\lfloor \frac{N-m}{k} \rfloor k}
\label{Higuchi_series}
\end{equation}
 , where $m = 1, 2, \cdots, k$.

For each time series constructed from (\ref{Higuchi_series}), the length $L(m,k)$ is computed by 
$$L(m,k) = \frac{\sum_{i=2}^{\lfloor \frac{N-m}{k} \rfloor}|x_{m+ik}-x_{m+(i-1)k}| (N-1)}{\lfloor \frac{N-m}{k} \rfloor k}$$

The average length $L(k)$ is computed as 
$$L(k) = \frac{\sum_{i=1}^{k}L(i,k)}{k}$$

This procedure repeats $k_{max}$ times for each $k$ from 1 to $k_{max}$, and then uses a least-square method to determine the slope of the line that best fits the curve of  $\ln(L(k))$ versus $\ln(1/k)$. The slope is the Higuchi Fractal Dimension. In this paper, $k_{max}=5$.

\subsection{Hjorth Parameters}
To a time series $x_{1}, x_{2}, \cdots, x_{N}$, the Hjorth mobility and complexity \cite{889990} are respectively defined as  $$\sqrt{\frac{M2}{TP}} \text{ and }\sqrt{\frac{M4\cdot TP} {M2 \cdot M2}}$$ , where $TP = \sum x_{i}/N$,  $\ M2 = \sum d_{i}/N$,  $M4 = \sum(d_{i}-d_{i-1})^{2}/N$ and $d_{i}=x_{i}-x_{i-1}$.

\section{Probabilistic Neural Network}
\label{PNN}
In machine learning, a classifier is essentially a mapping from the feature space to the class space. An Artificial Neural Network (ANN) implements such a mapping by using a group of  interconnected artificial neurons simulating human brain. An ANN can be trained to achieve expected classification results against the input and output information stream, such that there is not a need to provide a specified classification algorithm. 

PNN  is one kind of distance-based ANNs, using a bell-shape activation function. 
Compared with traditional back-propagation (BP) neural network, PNN is considered more suitable to medical application since it uses Bayesian strategy, a process familiar to medical decision makers~\cite{MedPNN}. Decision boundaries of PNN can be modified in real-time as new data becomes available~\cite{PNN}. There is no need to train the network over the entire data set again. We can therefore quickly update our network as more and more patients' data becomes available.

\setlength{\unitlength}{0.5cm}
 \begin{figure}[!htb]
\begin{center}
 \begin{picture}(20,9.5)
\put(1,0){\oval(1.5,1)[b]}
\put(1,8){\oval(1.5,1)[t]}
 \put(-0.5,9){Input Layer}
 \put(1,1.5){\makebox(0,0){\small{\textit{1$\times$R}}}}
 \put(1,3){\makebox(0,2){\rule{2mm}{20mm}}}
 \put(1,5){\vector(1,0){2}}
 \put(2,5){\makebox(0,0.6){\textbf{p}}}
 \put(2,5){\makebox(0,-0.7){\small{\small{\textit{1}}$\times$\textit{R}}}}
\put(6,0){\oval(8,1)[b]}
\put(6,8){\oval(8,1)[t]}
\put(4,9){Radial Basis Layer}
 \put(2.3,1){\makebox(0,0){1}}
 \put(2.5,1){\vector(1,0){1}}
 \put(3.5,0.5){\framebox(2,1){\textbf{b}}}
 \put(3.5,7){\framebox(2,1){$ \mathbf{W} $}}
 \put(3.5,6.5){\makebox(0,0){\small{\textit{Q}}$\times$\small{\textit{R}}}}
\put(3,4.5){\framebox(3,1){$\| \mathbf{W} - \mathbf{p}\|$}}
\put(4.5,7){\vector(0,-1){1.5}}
\put(4.5,0){\makebox(0,0){\small{\textit{Q}}$\times$\small{\textit{1}}}}
\put(4.5,3){\circle{1}}
\put(4.4,3){\makebox(0,0){\large{ $\cdot \times$}}}
\put(4.5,4.5){\vector(0,-1){1}}
\put(4.5,1.5){\vector(0,1){1}}
\put(5,3){\vector(1,0){2}}
\put(6,3.5){\makebox(0,0){$\mathbf{n}$}}
\put(6,2.5){\makebox(0,0){\small{\textit{Q}}$\times$1}}
\put(7,1){\framebox(2,4){\epsfig{file=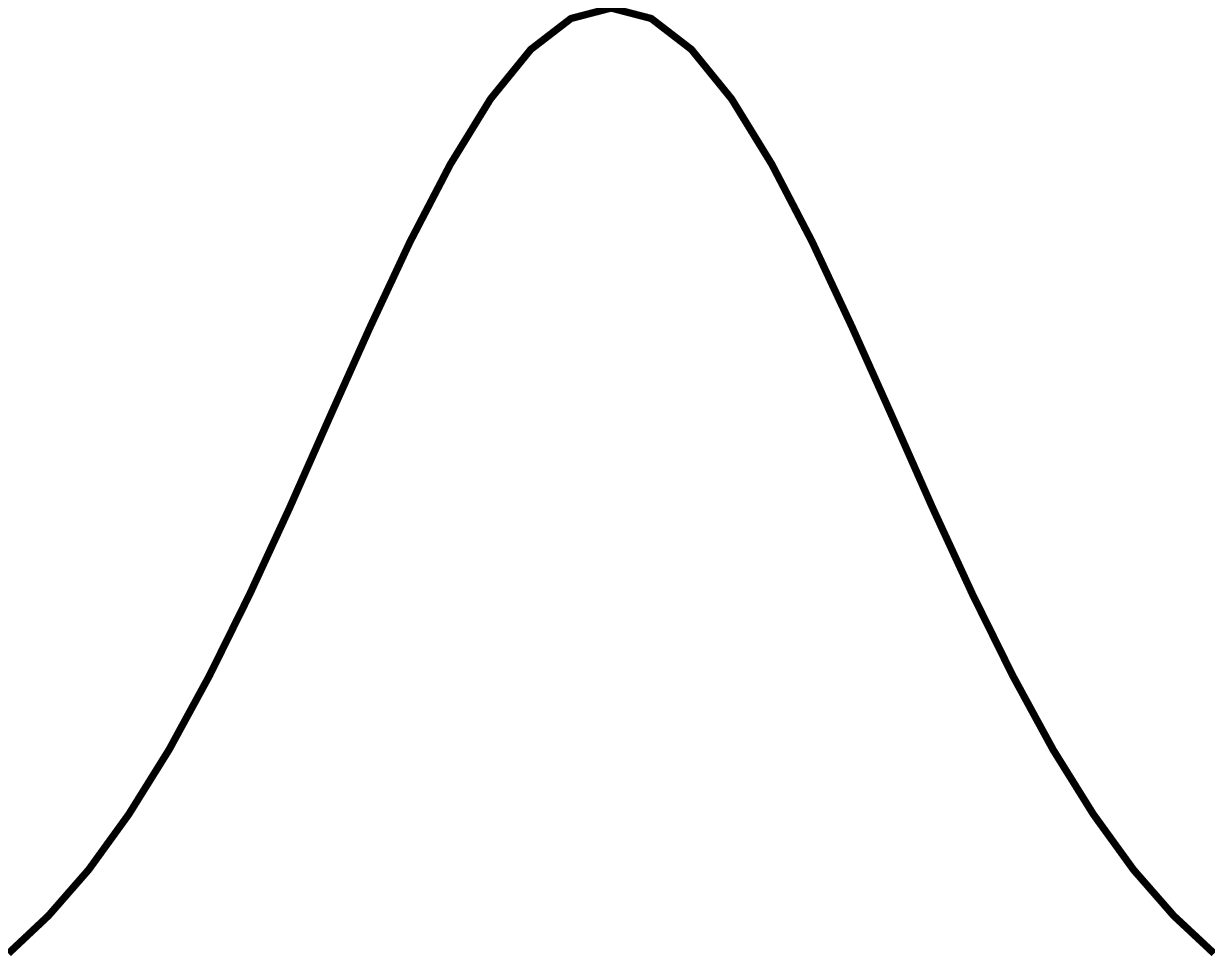,width=29pt,height=30pt}}}
\put(8,1.9){\makebox(0,-0.5){radbas}}
\put(8,0.5){\makebox(0,0){\small{\textit{Q}}}}
\put(9,3){\vector(1,0){2}}
\put(10,3.5){\makebox(0,0){$\mathbf{a}$}}
\put(10,2.5){\makebox(0,0){\small{\textit{Q}}$\times$1}}
\put(11,2.5){\framebox(2,1){$ \mathbf{M} $}}
\put(11.2,1.9){\small{\textit{K}}$\times$\small{\textit{Q}}}
\put(13,3){\vector(1,0){2}}
\put(14,3.5){\makebox(0,0){$\mathbf{d}$}}
\put(14,2.5){\makebox(0,0){\small{\textit{K}}$\times$1}}
\put(15,1){\framebox(1,4){\Large{C}}}
\put(15.5,0.5){\makebox(0,0){\small{\textit{K}}}}
\put(16,3){\vector(1,0){2}}
\put(17,3.5){\makebox(0,0){$\mathbf{c}$}}
\put(17,2.5){\makebox(0,0){\small{\textit{K}}$\times$1}}
\put(14,0){\oval(7,1)[b]}
\put(14,8){\oval(7,1)[t]}
\put(11,9){Competitive Layer}
\end{picture}
\caption{PNN structure, R: number of features, Q: number of training samples, K: number of classes. The input vector $\mathbf{p}$ is presented as a black vertical bar.}
\label{structure}
\end{center}
\end{figure}

Our PNN has three layers: the Input Layer, the Radial Basis Layer which evaluates distances between input vector and rows in weight matrix, and the Competitive Layer which determines the class with maximum probability to be correct. The network structure is illustrated in Fig. \ref{structure}. 
 Dimensions of matrices are marked under their names. 


\subsection{Radial Basis Layer}
In Radial Basis Layer, the vector distances between input vector $\mathbf{p}$ and the weight vector made of each row of weight matrix $\mathbf{W}$ are calculated. Here, the vector distance is defined as the dot product between two vectors\cite{PNN}. The dot product between $\mathbf{p}$ and the $i$-th row of $\mathbf{W}$ produces the $i$-th element of the distance vector matrix, denoted as $||\mathbf{W}  - \mathbf{p} ||$. The bias vector $\mathbf{b}$ is then combined with $||\mathbf{W} - \mathbf{p}||$ by an element-by-element multiplication, represented as ``$\cdot \times$" in Fig. \ref{structure}. The result is denoted as $\mathbf{n} = ||\mathbf{W} - \mathbf{p}|| \cdot \times \mathbf{b}$.

The transfer function in PNN has built into a distance criterion with respect to a center. In this paper, we define it as
\begin{equation}
 \text{radbas}(n) = e ^{-{n^2}}
\label{radbas}
\end{equation}
Each element of $\mathbf{n}$ is substituted into (\ref{radbas}) and produces corresponding element of $\mathbf{a}$, the output vector of Radial Basis Layer. We can represent the \textit{i}-th element of $\mathbf{a}$ as 
\begin{equation}
 \mathbf{a}_{i}=\text{radbas}(||\mathbf{W}_{i} - \mathbf{p}||\cdot \times \mathbf{b}_{i})
\end{equation}
, where $\mathbf{W}_{i}$ is the $i$-th row of $\mathbf{W}$ and $\mathbf{b}_{i}$ is the $i$-th element of bias vector $\mathbf{b}$.

\subsubsection{Radial Basis Layer Weights} Each row of $\mathbf{W}$ is the feature vector of one trainging sample. The number of rows equals to the number of training samples.

\subsubsection{Radial Basis Layer Biases} All biases in radial basis layer are set to $\sqrt{\ln{0.5}}/s$ resulting in radial basis functions that cross 0.5 at weighted inputs of $\pm s$, where $s$ is the spread constant of PNN. According to our experience,  $s=0.1$ can result in the highest accuracy.

\subsection{Competitive Layer}
There is no bias in Competitive Layer. In this layer, the vector $\mathbf{a}$ is first multiplied by layer weight matrix $\mathbf{M}$, producing an output vector $\mathbf{d}$. The competitive function $\mathbf{C}$ produces a 1 corresponding to the largest element of $\mathbf{d}$, and 0's elsewhere. The index of the 1 is the class of the EEG segment.
$\mathbf{M}$ is set to $K \times Q$ matrix of $Q$ target class vectors. If the $i$-th sample in training set is of class $j$, then we have a 1 on the $j$-th row of $i$-th column of  $\mathbf{M}$.

\section{Combining Classifiers Using Voting}
\label{bagging}
\begin{figure}[!hbt]
\begin{center}
\includegraphics[scale=0.6]{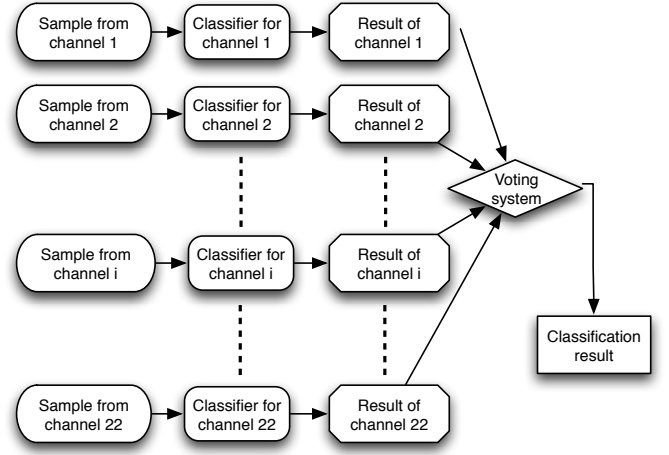}
\caption[fig]{Classification Voting Scheme}
\label{boosting}
\end{center}
\end{figure}

A simple voting scheme~\cite{Pattern_Classification} is used to improve the classification accuracy in this paper. We first build one  component classifier for each channel and then combine them as follows. Given 22 segments collected at the same time (from different channels), each of them will be classified by the component classifier for the same channel. The component classifier of each channel will judge whether the given EEG segment is epileptic. The final classification decision is based on the vote of each component classifier. The voting rule we use here is the majority rule.  Fig.~\ref{boosting} shows the diagram on how the combined classifiers work.

\section{Experimental Results}
\label{result}

In the experiments, we use MATLAB Neural Network Toolbox to implement PNN. The data used in the experiments is labeled as interictal (positive) or healthy (negative). The interictal data set has the same size as the healthy one.  The testing method for PNN is Leave-One-Out Cross-Validation (LOOCV)~\cite{Pattern_Classification}, where exactly one sample is used as the test sample while all the rest as training samples and such process repeats until every sample has been used as a test sample for exactly once.

We notice that different parameters used in feature extraction can lead to different classifier performance. We will show the experimental results using default feature extraction parameters in the first section while using optimized parameters in the second section.

\subsection{Classification using default feature extraction parameters}
The features are extracted using the default parameters described in Sec.~\ref{data}. We have carried out experiments to find the best features to be used for classification. We use all possible combinations of these features to build the PNN classifier: RIRs, Fractal Dimension (FDs) and Hjorth parameters (Hjorth's). The performance of each PNN with a specific combination of features is tested using LOOCV against each channel. The results are listed in Table~\ref{loocv} where each entry is the accuracy of LOOCV of the PNN with the features for that column against the data set of the channel corresponding to that row.

From Table~\ref{loocv}, it is clear that the first feature combination (using all features) yields the highest accuracy, and thus we decide to use all extracted features in later experiments to build the classifiers.

\begin{table}
\begin{center}
\caption[acc]{Single Channel Classification Accuracy Using PNN \label{loocv}}
\begin{tabular}[!htb]{@{\extracolsep{-2pt}} p{0.55cm} c c c c c c c }
\hline \multirow{2}{*}{channel}  & RIRs, FDs & FDs \& & \multirow{2}{*}{FDs} & \multirow{2}{*}{RIRs} & \multirow{2}{*}{Hjorth's} & RIRs &  RIRs \&  \\ 
&  \&  Hjorth's & Hjorth's & & & & \& FDs & Hjorth's  \\
\hline   Fp1 & 76 & 63 & 58 & 72 & 63 & 75 & 73 \\
\hline   Fp2 & 78 & 62 & 58 & 73 & 54 & 77 & 74 \\
\hline   F3 & 75 & 61 & 56 & 71 & 59 & 73 & 73 \\
\hline   F4 & 80 & 64 & 59 & 76 & 62 & 79 & 77 \\
\hline   C3 & 81 & 67 & 62 & 77 & 58 & 80 & 78 \\
\hline   C4 & 77 & 63 & 58 & 73 & 59 & 76 & 74 \\
\hline   P3 & 76 & 62 & 55 & 73 & 57 & 75 & 74 \\
\hline   P4 & 81 & 64 & 60 & 77 & 59 & 80 & 78 \\
\hline   O1 & 79 & 62 & 55 & 76 & 58 & 78 & 76 \\
\hline  O2 & 81 & 61 & 56 & 75 & 56 & 78 & 79 \\
\hline  F7 & 80 & 66 & 57 & 76 & 63 & 79 & 78 \\
\hline F8 & 85 & 70 & 57 & 81 & 61 & 82 & 84 \\
\hline T3 & 81 & 67 & 66 & 76 & 59 & 78 & 79 \\
\hline T4 & 81 & 62 & 60 & 78 & 53 & 80 & 79 \\
\hline T5 & 79 & 67 & 59 & 72 & 59 & 75 & 77 \\
\hline T6 & 78 & 67 & 57 & 70 & 62 & 74 & 75 \\
\hline A1 & 80 & 66 & 58 & 72 & 61 & 77 & 77 \\
\hline A2 & 80 & 61 & 56 & 72 & 60 & 76 & 75 \\
\hline Fz & 81 & 65 & 59 & 78 & 54 & 80 & 79 \\
\hline Pz & 79 & 65 &57 & 73 & 56 & 77 & 75 \\
\hline Cz & 81 & 66 & 62 & 77 & 56 & 80 & 78 \\
\hline Oz & 82 & 61 & 59 & 77 & 54 & 80 & 79\\
\hline 
\end{tabular}
\end{center}
\end{table}

The accuracy of the combined classifiers increases to 84.27\% while the true and false positive rates increase to 85.36\% and 83.18\% respectively. Thus, the sensitivity and specificity are 83.33\% and 84.69\%, respectively.



\subsection{Optimizing feature extraction parameters}
In Sec. \ref{data} and Sec. \ref{feature_extraction}, there are some parameters that can be changed: the segment length of EEG signal, the cut-off frequency of filters, and thebin($f_\text{step}$) and band ($f_\text{low}$ and $f_\text{up}$) in Eq. \eqref{PSI}. A combination of those parameters is called a \textit{configuration}. In this subsection, we will show that such configuration is important to the classification. Optimized configuration can lead to better accuracy. Different feature extraction parameters used in this paper are listed in Table \ref{para}. 

\begin{table}
\begin{center}
\caption[acc]{Feature Extraction Parameters Used In This Paper \label{para}}
\begin{tabular}[!hbt]{ c c }
\hline Parameters & Values\\ 
\hline segment length & 4096 or 8192 samples\\
\hline cut-off frequency of filters & 40, 46, 56 or 66 Hz\\
\hline  & band: 2-32 Hz, bin:1 Hz \\
spectral band and bin & band: 2-34 Hz, bin: 2 Hz\\
& band: 2-34.5 Hz, bin: 2.5 Hz\\
\hline
\end{tabular}
\end{center}
\end{table}

Table \ref{length} shows accuracies of combined PNN based classifiers in different configurations. The cut-off frequency of 56 and 66 Hz are not tested for segment length 4096, because we find longer segmentation can give higher accuracy. An interesting finding is that after the filter cut-off frequency reaches 46Hz, the accuracy does not significantly increase. One possible explanation is that many spikes may exist in interictal EEG and most reside in a frequency range of 15 to 50 Hz. Increasing the filter cut-off frequency may also introduce line noise from power supply or other sources, which will not benefit EEG signal quality~\cite{How_well_can_epileptic_seizures_be_predicted}. Table V shows the highest accuracy is 94.07\%, which is almost the same as the reported epilepsy diagnosis accuracy by human in a medical journal~\cite{EEG_accuracy}. 

\begin{table}
\begin{center}
\caption[acc]{Accuracy of Voted Classifier (PNN) in Different Configurations\label{length}}
\begin{tabular}[!hbt]{c| c |c c c }
\hline  \multirow{2}{*}{Length} & \multirow{2}{*}{cut-off freq.} & \multicolumn{3}{|c}{band and bin ($f_\text{low}$-$f_\text{up}$, $f_\text{step}$)}\\ 
 \cline{3-5} & & 2-32, 1 & 2-34, 2 & 2-34.5, 2.5\\
\hline  \multirow{2}{*}{4096} & 40 & 86.41 & 84.27 & 83.41 \\
\cline{2-5}  &46 & 91.77 & 89.81 & 89.23 \\
\hline  \multirow{4}{*}{8192}& 40 & 90.19 & 87.80 & 86.86\\
\cline{2-5} & 46 & 93.73 & 91.93 & 91.92 \\
\cline{2-5}  & 56 & \textit{94.07} & 92.14 & 91.37 \\
\cline{2-5} & 66 & 93.78 & 91.96 & 91.13  \\ \hline
\end{tabular}
\end{center}
\end{table}

\section{Conclusions}
In this paper, an automated interictal scalp EEG recognition system for epilepsy diagnosis is developed and validated. Three classes of features are extracted and PNNs are employed to make classification using those features. To improve the accuracy, we optimize the feature extraction parameters and design a final classifier that combines several PNN-based classifiers. Our system can reach an accuracy of 94.07\%, which is very close to the accuracy achieve by human. Compared with the existing approaches on epilepsy diagnosis, our approach does not require the occurrence of seizure activity during EEG recording. This merit reduces the difficulties in EEG collection since interictal data is much easier to be collected than ictal data. Therefore, our system is very helpful for areas short of medical resources.

\bibliographystyle{IEEEtran}
\bibliography{embc2009}

\end{document}